\documentclass[runningheads]{llncs}
\usepackage{graphicx}

\usepackage{tikz}
\usepackage{comment}
\usepackage{amsmath,amssymb} 
\usepackage{color}

\usepackage[accsupp]{axessibility}  

\usepackage{wrapfig}

\DeclareMathOperator*{\argmax}{arg\,max}

\usepackage[capitalize]{cleveref}
\crefname{section}{Sec.}{Secs.}
\Crefname{section}{Section}{Sections}
\Crefname{table}{Table}{Tables}
\crefname{table}{Tab.}{Tabs.}
\newcommand{\ie}{\textit{i.e., }}
\newcommand{\eg}{\textit{e.g., }}

\newcommand{\myparagraph}[1]{\vspace{2pt}\noindent{\bf #1}}

\usepackage{amsmath}
\usepackage{amssymb}
\usepackage{booktabs}
\usepackage{multirow}
\usepackage{adjustbox}
\usepackage{amssymb}
\usepackage{color}
\usepackage{booktabs}
\usepackage{lipsum}  
\usepackage{wrapfig}
\definecolor{battleshipgrey}{rgb}{0.52, 0.52, 0.51}

\usepackage{caption}
\usepackage{subcaption}

\begin{document}
\pagestyle{headings}
\mainmatter
\def\ECCVSubNumber{xxxx}  

\title{3D Compositional Zero-shot Learning with DeCompositional Consensus} 

\author{
Muhammad Ferjad Naeem$^*$\inst{1} \and Evin P{\i}nar \"{O}rnek$^*$\inst{2} \and Yongqin Xian\inst{1} \and Luc Van Gool\inst{1} \and Federico Tombari\inst{2,3}
}
\institute{ETH Zürich \and TUM \and Google}
\authorrunning{Naeem et al.}

\maketitle
\footnotetext[1]{First and second author contributed equally.}
\begin{abstract}
Parts represent a basic unit of geometric and semantic similarity across different objects. We argue that part knowledge should be composable beyond the observed object classes. Towards this, we present 3D Compositional Zero-shot Learning as a problem of part generalization from seen to unseen object classes for semantic segmentation. We provide a structured study through benchmarking the task with the proposed Compositional-PartNet dataset. This dataset is created by processing the original PartNet to maximize part overlap across different objects. The existing point cloud part segmentation methods fail to generalize to unseen object classes in this setting. As a solution, we propose DeCompositional Consensus, which combines a part segmentation network with a part scoring network. The key intuition to our approach is that a segmentation mask over some parts should have a consensus with its part scores when each part is taken apart. The two networks reason over different part combinations defined in a per-object part prior to generate the most suitable segmentation mask. We demonstrate that our method allows compositional zero-shot segmentation and generalized zero-shot classification, and establishes the state of the art on both tasks.
\end{abstract}

\section{Introduction}
\label{sec:intro}

A centaur is a mythological creature with the upper body of a human and the bottom body of a horse. This creature was never observed in our world, yet even a child can label its body parts from the human head to the horse legs. We humans can dissect the knowledge of basic concepts as primitives, like parts from human head to horse legs, to generalize to unseen objects. Cognitive studies have shown that humans learn part-whole relations in hippocampal memory to achieve object understanding through compositionality \cite{HINTON1979231,ROLLS1994335}. Compositionality has evolved as a survival need since every combination of every primitive cannot be observed.

\begin{figure}[t]
  \centering
  \includegraphics[width=1\linewidth]{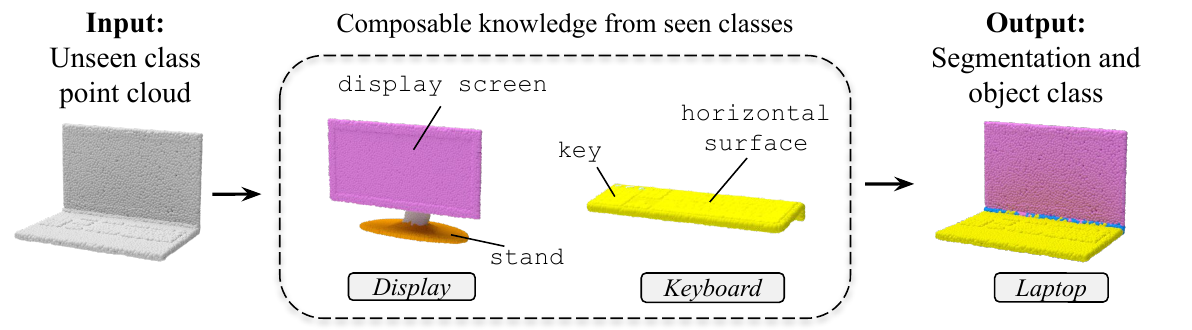}
   \caption{We aim to compose parts (\eg \texttt{display screen, key, horizontal surface)} from seen (\eg \textit{Display, Keyboard}) to unseen object classes (\eg \textit{Laptop}) for semantic segmentation and classification in 3D point clouds.
   }
   \label{fig:teaser}
\end{figure}

Parts represent a basic primitive of geometric and semantic similarity across objects. Recently, PartNet dataset has been introduced to study fine-grained semantic segmentation of parts \cite{Mo_2019_CVPR}. This has inspired several architectural works towards improving supervised fine-grained segmentation in 3D models \cite{dgcnn,convpoint,gdanet}. 
A parallel line of work uses the concept of parts to improve tasks like 3D reconstruction with hierarchical decomposition \cite{Paschalidou2020CVPR}, unsupervised segmentation by finding repeated structural patterns \cite{luo2020learning}, and instance segmentation in unseen objects \cite{chen2019bae_net}. However, these works do not predict semantic part classes.

With the availability of RGB-D sensors and the ease of acquiring 3D data in domains from augmented reality to robotic perception, the need for object understanding beyond seen object classes has emerged \cite{kinectfusion,3dsis,Wald2019RIO}. A model is unlikely to be trained for all possible existing objects \cite{Cheraghian2020TransductiveZL,cheraghian2019ZSL3D}, however, man-made environments consist of objects that share similarities through their parts. In this scenario, reasoning over learned parts can present an avenue for generalization to unseen objects classes. 
Zero-shot learning with 3D data has received far less attention compared to 2D domain. 
In this work, we introduce a new task, namely 3D Compositional Zero-Shot Learning (3D-CZSL), aiming at jointly segmenting and classifying 3D point clouds of both seen and unseen object classes (see Figure \ref{fig:teaser}). 3D-CZSL is a challenging task as it requires generalizing parts from seen object classes to unseen classes that can be composed entirely of these parts.

Our contributions are as follows: 
(1) We formalize zero-shot compositionality for 3D object understanding with semantic parts and introduce the 3D-CZSL task.
To the best of our knowledge, we present the first work for joint classification and semantic part labeling for compositional zero-shot learning in 3D.
(2) We establish a novel benchmark through Compositional PartNet (C-PartNet), which enables research in 3D-CZSL through 16 seen and 8 unseen object classes. 
(3) We show that existing point cloud models fail to generalize beyond the seen object classes, whereas the performance of existing 2D zero-shot methods is severely limited in the 3D domain.
(4) We propose a novel method, DeCompositional Consensus, which maximizes agreement between a segmentation hypothesis and its decomposed parts. Our method sets the state of the art for 3D-CZSL.


\section{Related Work}

Our work lies at the intersection of compositionality, zero-shot learning, and 3D point cloud part segmentation and discovery. 

\myparagraph{Compositionality} is the notion of describing a whole through its parts, studied thoroughly in many disciplines such as mathematics, physics, and linguistics. Hoffman \cite{hoffman1984parts} and Biederman \cite{biederman1987} suggested that human object recognition is based on compositionality. They heavily influenced both traditional and modern computer vision research, such as describing objects by their primitives in Deformable Part Models \cite{dpm2008}, images as a hierarchy of features in Convolutional Neural Networks \cite{zeiler2014visualizing,lecun1989backpropagation}, understanding a scene through its components as in Scene Graphs \cite{johnson2015image}, events as a set of actions as in Space-Time Region Graphs \cite{Wang_videogcnECCV2018}. Parts have been used as semantic and geometric object primitives, which were seen to be captured within CNN kernels implicitly \cite{garcia2018,garcia2018_2}.

\myparagraph{Zero-shot learning (ZSL)} addresses the task of recognizing object classes whose instances have not been seen during training \cite{larochelle2008,socher2013,thegoodthebad}. This is attained through auxiliary information in the form of attributes (ALE \cite{ale}), word embeddings (SPNet \cite{spnet}), or text descriptions \cite{RALS16}. 
\textit{\textbf{Compositional zero-shot learning (CZSL)}} focuses on detecting unseen compositions of already observed primitives. The current literature on the topic focuses on state-object compositionality. Towards this, one line of research aims to learn a transformation between objects and states \cite{redwine,aopp,symnet}. Another line proposes a joint compatibility function with respect to the image, state, and object \cite{tmn,yang2020learning,compcos}. Graph methods are also recently used in this direction including learning a causal graph of state object transformations \cite{causal} and using the dependency structure of state object compositions to learn graph embeddings \cite{cge,cocge}. There have been some preliminary works exploring \textit{\textbf{zero-shot learning in 3D}} as an extension of 2D methods including projecting on word embeddings \cite{cheraghian2019ZSL3D}, using transductive approaches \cite{Cheraghian2020TransductiveZL}, along with some unlabelled data \cite{CheraghianRCP19}, and using generative models to learn the label distribution of unseen classes \cite{bjorn2021}. 

\myparagraph{3D part segmentation and discovery} aims at parsing 3D objects into semantically and geometrically significant parts. The PartNet dataset \cite{Mo_2019_CVPR} enabled studying fine-grained 3D semantic segmentation, hierarchical segmentation, and instance segmentation. The existing point cloud processing methods accomplish the task through conditioning the model over the known object class. PointNet models \cite{qi2016pointnet,qi2017pointnetplusplus} provide multi-layer-perceptron (MLP) based solutions, DGCNN \cite{dgcnn} uses graph convolutions for point clouds, ConvPoint \cite{convpoint} pre-processes points to define neighborhoods for convolutions, GDANet \cite{gdanet} uses attention in addition to MLP and currently holds state of the art for part semantic segmentation. Capsule Networks \cite{capsuledynamic,capsnet} propose architectural changes that implicitly model parts for tasks like object classification and segmentation.

An alternate line of work uses the idea of parts in objects for downstream tasks like instance segmentation and point cloud reconstruction. This includes discovering geometrically similar part prototypes, similar to superpixels \cite{luo2020learning}, predicting category-agnostic segmentation through a clustering approach \cite{Wang_2021_CVPR},
finding repetitive structural patterns in instances of an object \cite{chen2019bae_net}, modeling 3D objects as compositions of cuboids \cite{abstractionTulsiani17}, superquadrics \cite{superquadrics,Paschalidou2019CVPR,Paschalidou2020CVPR}, 
convex functions \cite{deng2020cvxnet}, and binary space partitioning planes \cite{chen2020bspnet} through deep learning. 

Our work lies at the intersection of these three areas. Similar to CZSL works \cite{redwine,aopp,symnet,cge,causal}, we study the compositionality of learned primitives, however, we are interested in parts of objects rather than state-object relations. Similar to ZSL works \cite{larochelle2008,socher2013,thegoodthebad,ale,spnet,RALS16}, we learn classification scores of unseen object classes, however, our method only uses parts as side information and does not rely on any pretrained models like word embeddings. Similar to part discovery in objects \cite{luo2020learning,chen2019bae_net,Paschalidou2020CVPR,kawana2021unsupervised,capsnet}, we rely on parts as a basic unit of understanding an object. However, instead of geometric primitives, we use human-defined semantic parts, which tightly couple geometry, semantics, and affordances \cite{deng20213d}. 
Our method further has parallels to ensemble learning, where a combination of learners solves the same downstream task \cite{ensemble_zeroshot}, however, we use an agreement between different tasks to improve generalization.

\begin{figure*}[t]
  \centering

   \includegraphics[width=\linewidth]{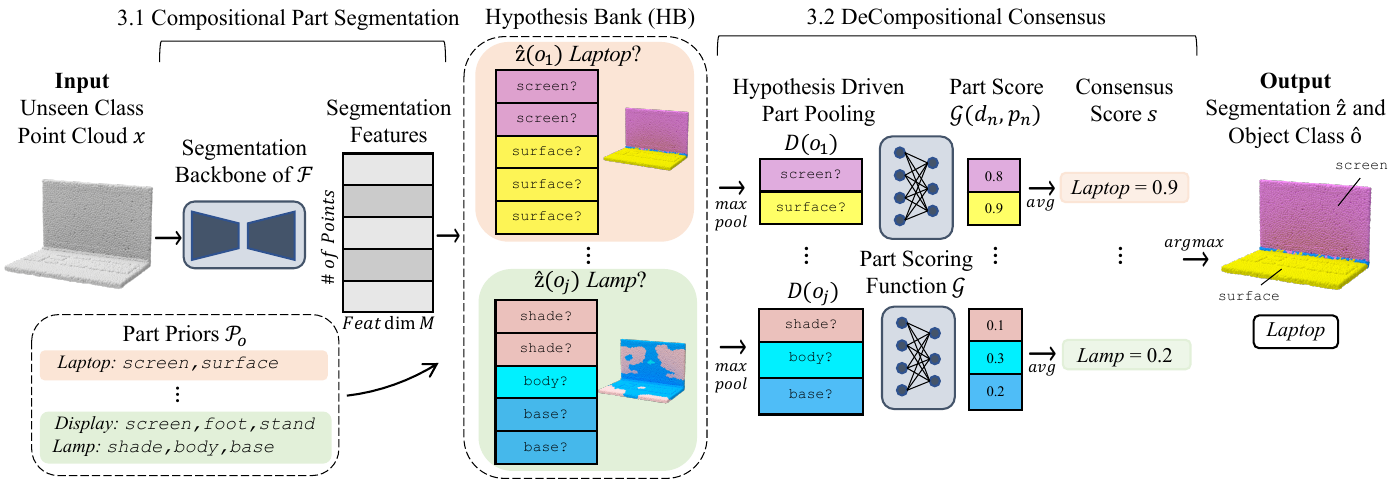}
   \caption{
   \textbf{DeCompositional Consensus(DCC)} combines our compositional part segmentation function $\mathcal{F}$ with our part scoring function $\mathcal{G}$. We use the Part Prior $\mathcal{P}_o$ of what parts can exist in each object class to populate the Hypothesis Bank with multiple segmentation masks. These hypotheses are used in a Hypothesis Driven Part Pooling to get a part descriptor of each part as an input to the part scoring function $\mathcal{G}$ to calculate the DCC Score.
   This score measures the agreement of the segmentation mask with its part scores when each part is taken apart like lego blocks.
   Hypothesis with the maximum DCC score is selected for compositional zero-shot segmentation and zero-shot classification.
   }
   \label{fig:arch}
\end{figure*}

\section{Proposed Approach}
\label{sec:approach}

In the following, we formalize the problem and explain the proposed solution.

\myparagraph{Problem formulation.} 
Let $\mathcal{T}$ define the training set with instances $(x, o, z)$, where $x$ is an input object point cloud described as a set of points in $\mathbb{R}^{3}$, $o$ is the object class label from the set of seen object classes $\mathcal{O}_s$ and $z$ is the part segmentation mask labelled with parts $p$ from the set of all possible parts $\mathcal{P}$. We task a model to generalize to a set of unseen object classes $\mathcal{O}_u$, \ie $\mathcal{O}_s\cap \mathcal{O}_u=\emptyset$.
We assume that $\mathcal{O}_u$ is labelled with the same part set $P$ for part segmentation that was completely observed in seen object classes $\mathcal{O}_s.$ This makes the part segmentation task as a compositional zero-shot problem and the object classification task as a generalized zero-shot problem, \ie we predict over the full object set $\mathcal{O} = \mathcal{O}_s \cup \mathcal{O}_u$ at inference for object classification.
We further assume that the model has access to a \textbf{part prior} for all object classes. For an object class $o$, this prior is defined as the set of parts $\mathcal{P}_{o}=\{p_1,...,p_l\}$ that it can be labelled with for part segmentation. 

\myparagraph{Method overview.} 
Part segmentation is a challenging task, as it requires one model to adapt to parts of varying scale, orientation, and geometry for all objects. Existing point cloud part segmentation methods simplify this by learning an object class conditioned model, either by training separate models specialized for each object class \cite{Mo_2019_CVPR}, or by feeding the object class label as an input to the model (one-hot class vector \cite{qi2016pointnet,dgcnn,gdanet,capsnet}). However, this requires an object class input at test time which is not available for unseen object classes. 
In this work, we refer to this case as \textbf{object prior}, \ie the model has access to the ground truth object class.
The first step of our approach removes the object prior assumption and proposes Compositional Part Segmentation. 
In the second step, we propose our model DeCompositional Consensus which predicts the object class using the part segmentation from the previous step. It learns an agreement over a segmentation hypothesis and its part-based object classification score based on the idea of an object being taken apart like Lego blocks. The full model is depicted in Figure \ref{fig:arch}.

\subsection{Compositional Part Segmentation}
\label{app:comppartseg}
We reformulate part segmentation to allow compositional reasoning by encoding the part prior into the optimization criterion and the model inference. 
Formally, given an input point cloud $x$, we define $\mathcal{F}(x, p)$ as the part segmentation function, with learnable parameters $W$, which returns a part score for each part $p$ in the full part set $\mathcal{P}$. 
At training time, we compute the segmentation loss $L_{Seg}$ from \cite{qi2016pointnet,qi2017pointnetplusplus} as a cross entropy over parts in the part prior $\mathcal{P}_o$ of the ground truth object class $o$ rather than the full part set $\mathcal{P}$. 
At inference, the predicted part segmentation mask $\hat{z}(o)$ of an object class $o$ is computed over the scores of parts in its part prior: 
\begin{equation}
\label{eq:inference}
\hat{z}(o) = \argmax_{p \in {\mathcal{P}_{o}}}(\mathcal{F}(x, p))
\end{equation}

With the proposed changes, a part segmentation model such as PointNet\cite{qi2016pointnet} can now compositionally generate a part segmentation mask for any object class we have a part prior for. 
Furthermore, the proposed improvements also prevent unintended biases against similar parts in different object classes as shown experimentally later in Table \ref{tab:bboracle}.
Notably, when an object prior is available as ground truth class, it defines the upper bound of the part segmentation performance for a model (see Table \ref{tab:basenumberseg} ``Object Prior"). 
In the absence of an object prior (GT object class), we can predict $|\mathcal{O}|$ segmentation masks (hypotheses) for each object class we have a part prior for, generating the \textbf{Hypothesis Bank (HB)}. Next, we introduce our novel method which allows for selecting the most suitable segmentation hypothesis for an input point cloud.
\subsection{DeCompositional Consensus}
\label{app:dcc}
We propose a novel method, \textbf{DeCompositional Consensus (DCC)}, 
which learns an agreement (Consensus) over a segmentation hypothesis and its part-based object classification score 
when the object is taken apart (DeComposed) into parts like lego blocks as segmented in the hypothesis. 
DCC is based on the idea that we can learn what a valid part descriptor is from seen object classes to generalize to unseen object classes. 

\myparagraph{Hypothesis driven part pooling.} We extract a part descriptor for each part in a segmentation hypothesis from the point-wise features of the segmentation backbone as shown in Figure \ref{fig:arch}.
Part segmentation models like PointNet\cite{qi2016pointnet} generate this feature representation in the penultimate layer of the model, \ie before the final per-part segmentation scoring layer.
We use the segmentation hypothesis as the pooling mask to pool over the point dimensions of the feature map for each part. This results in a permutation invariant part feature vector for each part, \ie part descriptor representing the features responsible for that part segmentation in this hypothesis. 
We choose maxpool as the pooling operation due to its wide adoption in point cloud literature \cite{qi2016pointnet,qi2017pointnetplusplus}. 
For the segmentation hypothesis $\hat{z}(o)$ of an object class $o$, this operation returns a set $\mathcal{D}(o)$ with part descriptors $d$ for each part $p$ found in this hypothesis. Note that $|\mathcal{D}(o)|$ is not always equal to $|\mathcal{P}_o|$ as a segmentation hypothesis might not contain all parts defined in the $\mathcal{P}_o$, \eg an instance of a chair might or might not contain sidearms.

\myparagraph{Learning DeCompositional Consensus.} 
Our DCC model learns a part scoring function $\mathcal{G}$ with weights $\Theta$. For a part descriptor $d$, the function returns a score $\mathcal{G}(d, p)$ which measures the likelihood of this part descriptor to belong to the part $p$. We define DeCompositional Consensus score as the agreement between the segmentation hypothesis and the part scores. For an object hypothesis $\hat{z}(o)$, the DCC score is defined as:
%
\begin{equation}
    \label{eq:hypscore}
s(x, o) = \frac{1}{|\mathcal{D}(o)|}{\sum_{n=1}^{|{\mathcal{D}(o)}|}    \mathcal{G}(d_n, p_n)}
\end{equation}
Our novel DCC score measures the individual consensus of each part descriptor with the full segmentation mask to define an object classification score. We optimize DCC score for classification with a cross entropy loss over $\mathcal{O}_s$ as:
\begin{equation}
\label{eq:decomp}
    L_{DeComp} = 
    -log ( \frac{\exp{s(x, o)}}
    {\sum_{o'\in \mathcal{O}_{s}} \exp{s(x, o')}}  )
\end{equation}

Since $L_{DeComp}$ is computed over the Hypothesis Bank generated by $\mathcal{F}$, an additional part classification loss $L_{Part}$ is computed using the ground truth segmentation mask $z$ of each input to prevent bias against parts that are hard to segment. $L_{Part}$ uses the ground truth segmentation mask to extract part descriptor set $\mathcal{D}_{gt}$ and optimizes them for part classification over $\mathcal{P}$.
\begin{equation}
\label{eq:partcls}
    L_{Part} = 
    \sum_{n=1}^{|{\mathcal{D}_{gt}}|} -log ( \frac{\exp{\mathcal{G}(d_n, p_n)}}
    {\sum_{p'\in \mathcal{P}} \exp{\mathcal{G}(d_n, p')}}  )
\end{equation}

\myparagraph{Inference.}
For generalized zero-shot inference, the HB is populated over all object classes $\mathcal{O} = \mathcal{O}_s + \mathcal{O}_u$.
The object class prediction $\hat{o}$ for an input point cloud $x$ is retrieved by selecting the object class with the highest DeCompositional Consensus score:
\begin{equation}
    \hat{o} = \argmax_{o' \in {\mathcal{O}}}(s(x, o')) 
\end{equation}
The corresponding hypothesis of the predicted class $\hat{o}$ becomes the final part segmentation output, \ie $\hat{z}(\hat{o})$.
Our technical novelty lies in defining part descriptors as features responsible for part segmentation in a hypothesis; and using their likelihood to define an object class level consensus score to achieve zero-shot compositionality. 
In contrast to several zero-shot baselines \cite{spnet,cge}, our method does not require any supervised calibration step over the unseen classes.


\section{Compositional PartNet Benchmark}

\begin{figure*}[t]
\centering
 \includegraphics[width=1.0\linewidth]{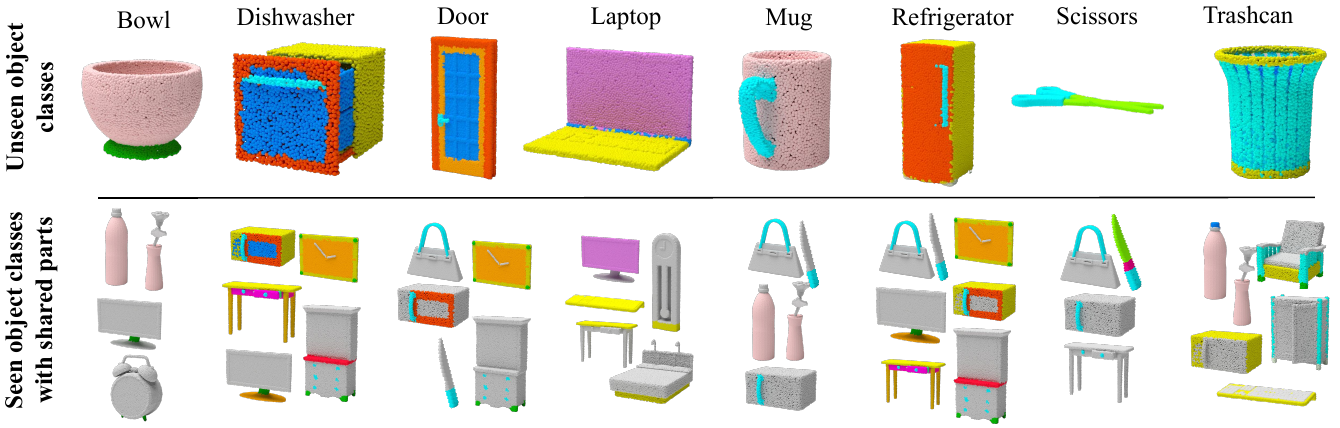}
 \caption{\textbf{Compositional PartNet} refines the labels of PartNet dataset to maximize shared parts across different object classes, and enables studying 3D-CZSL task. The available 24 object classes are divided into 16 seen classes for training and 8 unseen classes for inference in zero-shot. We depict the shared labels between seen and unseen object classes in same colors.}
 \label{fig:dataset}
\end{figure*}

Zero-shot compositionality in machine learning algorithms has mainly been studied for state-object relations in image datasets like MIT-States \cite{mitstates}, UT-Zappos \cite{utzappos},  AO-CLEVr \cite{causal}, and more recent C-GQA \cite{cge}. These datasets have several limitations such as including label noise \cite{mitstates,causal}, lacking visual cues \cite{utzappos,cge}, being too simple \cite{causal}, or missing multilabel information \cite{cge}. 

We believe that 3D part object relations provide an ideal avenue to study zero-shot compositionality, as they tend to be more well-defined albeit challenging.
There have been several attempts in a part-based benchmark \cite{chen2009,yi2016} for 3D object understanding. 
Recently, ShapeNet has been extended with fine-grained part labels to form the new dataset PartNet \cite{Mo_2019_CVPR}. 
PartNet provides 24 distinct object classes, annotated with fine-grained, instance-level, and hierarchical 3D part information, consisting of around 26K 3D models with over 500K part instances and 128 part classes. 
However, these part class labels are not unified across different object categories, preventing a study into zero-shot compositionality. 
We refine PartNet into \textbf{Compositional PartNet (C-PartNet)} with a new labeling scheme that relates the compositional knowledge between objects by merging and renaming the repeated labels as shown in Figure \ref{fig:dataset}. 

\myparagraph{Unifying part labels.} 
While PartNet provides three levels of hierarchical part labels, not all objects are labeled at the deepest level. We take the \textbf{deepest level} available for each object. We find similar parts within and across different objects by training a supervised segmentation model and compute pairwise similarities between parts across PartNet. Parts that share a high similarity and have the same semantic meaning (\eg \texttt{bed horizontal surface} in object Bed and \texttt{horizontal surface} in Storage Furniture) are merged into a single general part label (\texttt{horizontal surface}). Furthermore, parts with a similar function but different name (\eg \texttt{screen side} of Laptop and \texttt{display screen} of Display) are merged together. The relabelled C-PartNet consists of 96 parts compared to 128 distinct part labels in the original PartNet. Details in the supplementary.

\myparagraph{Selecting test time unseen object classes.}
Objects that share a similar function tend to have similar parts \cite{biederman1987}. We divide PartNet objects into several functional categories. Details of this categorization and the dataset statistics can be found in the supplementary. We identify three easy to compose unseen object classes (\ie Mug, Bowl and TrashCan), that share large similarities with seen object classes (Bottle and Vase). 
Furthermore, we choose three object classes of medium difficulty that require generalizing parts beyond the context they were observed in (\ie Dishwasher, Refrigerator, and Laptop). Finally, Scissors and Door present two hard-to-compose object classes that require generalizing beyond scale, context, and number instances of parts compared to seen object classes. The validation set contains all seen and 2 unseen object classes (Bowl and Dishwasher). The test set consists of 16 seen $O_s$ and 8 unseen classes $O_u$.


\section{Experiments}

Since our proposed benchmark lies at the intersection of point cloud processing, attribute learning, zero-shot learning, and its specialized sub-domain compositional zero-shot learning, we adapt baselines representing these lines of works.

\myparagraph{Baselines.}
\textit{Object Prior} uses a point cloud part segmentation model trained with our framework and evaluates the segmentation performance on the ground truth object. This is the oracle upper bound for the zero-shot models. \textit{Direct Seg} trains a point cloud part segmentation model $\mathcal{F}$ without a part prior to predict over all parts $\mathcal{P}$ in the dataset. \textit{PartPred} is inspired from a classic zero-shot baseline DAP \cite{attpred} and trains a part prediction network from the global feature of each point cloud. The predicted parts are used as $P_o$ for equation \ref{eq:inference} to condition the compositional part segmentation network \cite{attpred}. For zero-shot classification baselines, we use the predicted class to select the corresponding segmentation mask from the Hypothesis Bank. 
Among these, \textit{SPNet} \cite{spnet} learns classification by projecting the global feature of an input on a pretrained distribution where both seen and unseen objects lie \eg word embeddings. \textit{CGE} \cite{cge} proposes to model compositional relations using a graph consisting of parts connected to objects they occur in. We reformulate CGE to a multitask setup and use part nodes for segmentation and object nodes for classification. 
\textit{PartPred DCC} uses the part prediction network's scores for parts found in each segmentation hypothesis to calculate the consensus score from Equation \ref{eq:hypscore}.
Finally, \textit{3D Capsule Networks} \cite{capsnet} aim to discover part prototypes through unsupervised reconstruction. Segmentation is subsequently learned by a linear mapper from capsules to part labels. 
We give additional details about these baselines in the supplementary and also compare with the current SOTA for part class agnostic segmentation method, Learning to Group \cite{luo2020learning}, on unseen object classes. 


\myparagraph{Metrics.}
The proposed benchmark consists of two jointly learned tasks.
For the \textbf{compositional zero-shot segmentation}, we report the mean object class-wise Intersection-over-Union (mIoU) for part labels over seen and unseen object classes. 
We also report the harmonic mean over seen and unseen object classes to study the best generalized zero-shot performance. In addition, we report a per-object mIoU to study model performance on each unseen object across the three difficulty levels.
For \textbf{generalized zero-shot classification}, we report for the per-object class top-1 classification accuracy over unseen classes, mean accuracy over seen classes, unseen classes and their harmonic mean. For models that apply joint classification and segmentation, we choose the checkpoint with the best segmentation performance to encourage compositional part understanding. Part based classification baselines can give the same scores across two objects if an instance does not have all parts, \eg an empty Vase has the same parts in Vase Hypothesis and Bowl. This is counted as an accurate classification, since the ground truth object still receives the highest score and achieves compositional segmentation.

\myparagraph{Training details.}
For its simplicity and competitive performance in our ablations (see Table \ref{tab:backbone}), we choose PointNet \cite{qi2016pointnet} as the backbone model for $\mathcal{F}$ in our baseline comparisons in Table \ref{tab:basenumberseg}, \ref{tab:basenumbercls}. We also report further results on DGCNN \cite{dgcnn}, ConvPoint \cite{convpoint} and GDANet \cite{gdanet} in Table \ref{tab:backbone} . All backbones are pretrained with the author's implementations extended by our framework. The pretrained models are then used as initialization for the zero-shot models and are finetuned.
For our model DCC, we use a 2-layer MLP with $512$ hidden dimensions, ReLU, and dropout followed by a linear layer as function $\mathcal{G}$. We use a step size learning rate scheduler between $1e^{-3}$ and $1e^{-5}$ with Adam optimizer. We use cross entropy as segmentation loss $L_{Seg}$ for part segmentation similar to \cite{qi2016pointnet,qi2017pointnetplusplus,dgcnn,gdanet}. $L_{Seg}$, $L_{DeComp}$ and $L_{Part}$ are equally weighted and the network is trained until convergence on the validation set. We use Word2Vec \cite{word2vec} for models that rely on word embeddings \cite{spnet,cge}. For CGE, we choose the graph configuration that achieved the best result on the validation set at 2 layers of GCN with a hidden dimension of $1024$. 
Our framework is implemented in PyTorch \cite{paszke2019pytorch} and all experiments are conducted using Nvidia A100 GPUs. The dataset and experimental framework will be released upon acceptance.

{
\setlength{\tabcolsep}{2pt}
\renewcommand{\arraystretch}{1.3}
\begin{table}[t]
    \centering
    
    \begin{subtable}[t]{\linewidth}
    \resizebox{\linewidth}{!}
    {
    \begin{tabular}{l|ccc|cccccccc }
    \multirow{2}{*}{\bfseries Method}  & \multicolumn{3}{c}{\textbf{}} &
    \multicolumn{8}{c}{\textbf{Unseen object classes}}\\
     &  HM & S & U & 
    Bowl& Dish & Door &  Lap & Mug  & Refr & Scis & Trash \\
    \midrule
    Object Prior \cite{qi2016pointnet} & 
    47.9 & 52.8 & 43.8 &
    77.0 & 40.2 & 25.1 & 72.4 & 47.1 & 31.9 & 22.5 & 34.2 \\
    \midrule
    Direct Seg \cite{qi2016pointnet} &  
    28.5 & \textbf{48.7} & 20.1 &
    62.9 & 4.0 & 1.6 & 19.9 & 35.7 & 0.9 & 0.0 & 33.9 \\
    SPNet* \cite{spnet} &  
    8.5 & 28.5 & 5.0 &
    12.6 & 2.5 & 0.5 & 0.0 & 2.6 & 2.6 & 0.0 & 15.8 \\
    
    CGE* \cite{cge} &  
    30.8 & 37.0 & 26.4 &
    67.0 & 19.5 & 0.3 & 35.1 & 39.6 & 11.2 & 0.0 & 33.6 \\
    
     3D-PointCapsNet \cite{capsnet} & 4.4 & 9.4 & 2.9 & 4.3 & 0.0 & 0.2 & 1.2 & 11.2 & 0.1 & 0.1 & 6.5 \\
    
    PartPred \cite{attpred} & 
    26.3 & 33.6 & 21.6 &
    66.2 & 2.3 & {7.2} & 19.4 & 43.1 & 0.5 & 0.0 & 32.5 \\
    
    PartPred DCC \cite{attpred} & 
    20.9 & {41.3} & 14.0 &
    35.5 & 2.1 & \textbf{7.2} & 17.2 & 29.2 & 0.7 & 0.0 & 20.0 \\
    
    \hline
    
    DCC (ours) & \textbf{35.2} & 38.0 & \textbf{32.7} &
    \textbf{66.1} & \textbf{30.9} & 5.3 & \textbf{56.3} & \textbf{40.4} & \textbf{28.4} & 0.0 & \textbf{34.2} \\
    
    \end{tabular}}
    \caption{\textbf{Compositional Zero-shot Segmentation} }
    \label{tab:basenumberseg}
    \end{subtable}
    \hspace{\fill}
    \begin{subtable}[t]{\linewidth}
    \resizebox{\linewidth}{!}
    {
    \begin{tabular}{l|ccc|cccccccc }
    \multirow{2}{*}{\textbf{Method}} &  \multicolumn{3}{c}{\textbf{}} &
    \multicolumn{8}{c}{\textbf{Unseen object classes}} \\
    &   HM & S & U & 
    Bowl& Dish & Door &  Lap & Mug  & Refr & Scis & Trash \\
    \midrule
    SPNet* \cite{spnet} &  
    3.8 & 46.7 & 2.0 &
    12.0 & 0.0 & 3.1 & 0.0 & 0.0 & 0.0 & 0.0 & 0.0 \\
    CGE* \cite{cge} &  
    33.1 & 54.3 & 23.8 &
    31.9 & 0.0 & 0.0 & 52.0 & 1.0 & 33.7 & 0.0 & \textbf{71.9} \\
    PartPred DCC \cite{attpred}\hspace{13pt} & 
    19.9 & \textbf{74.0} & 11.5 &
    4.3 & 3.3 & \textbf{25.8} & 0.0 & 13.5 & 0.0 & 0.0 & 45.2 \\
    \hline
    
    DCC(ours) & \textbf{55.9} & {73.2} & \textbf{45.2} &
    \textbf{79.8} & \textbf{57.1} & 5.3 & \textbf{55.4} & \textbf{71.9} & \textbf{55.6} & 0.0 & 36.8 \\
    \end{tabular}}
    \caption{\textbf{Generalized Zero-shot Classification}}
    \label{tab:basenumbercls}
    \end{subtable}
    
    \caption{\textbf{Baseline comparison.} We compare our proposed method, DeCompositional Consensus (DCC), against baseline and report results for the two tasks. * marks baselines that require supervised calibration. For (a), we report mIoU \% over part labels per object class over seen objects, unseen objects, and their harmonic mean. We also report the mIoU over each unseen object class. For (b), we report the top-1 classification accuracy. DCC achieves SOTA on both tasks.}
\end{table}
}

\subsection{Comparing with State of the Art}
We compare our method with baselines on compositional zero-shot segmentation in Table \ref{tab:basenumberseg} and generalized zero-shot classification in Table \ref{tab:basenumbercls}. Our method outperforms all baselines on almost all metrics and establishes state of the art on both tasks. 

\begin{figure}[t]
  \centering
   \includegraphics[width=1.\linewidth]{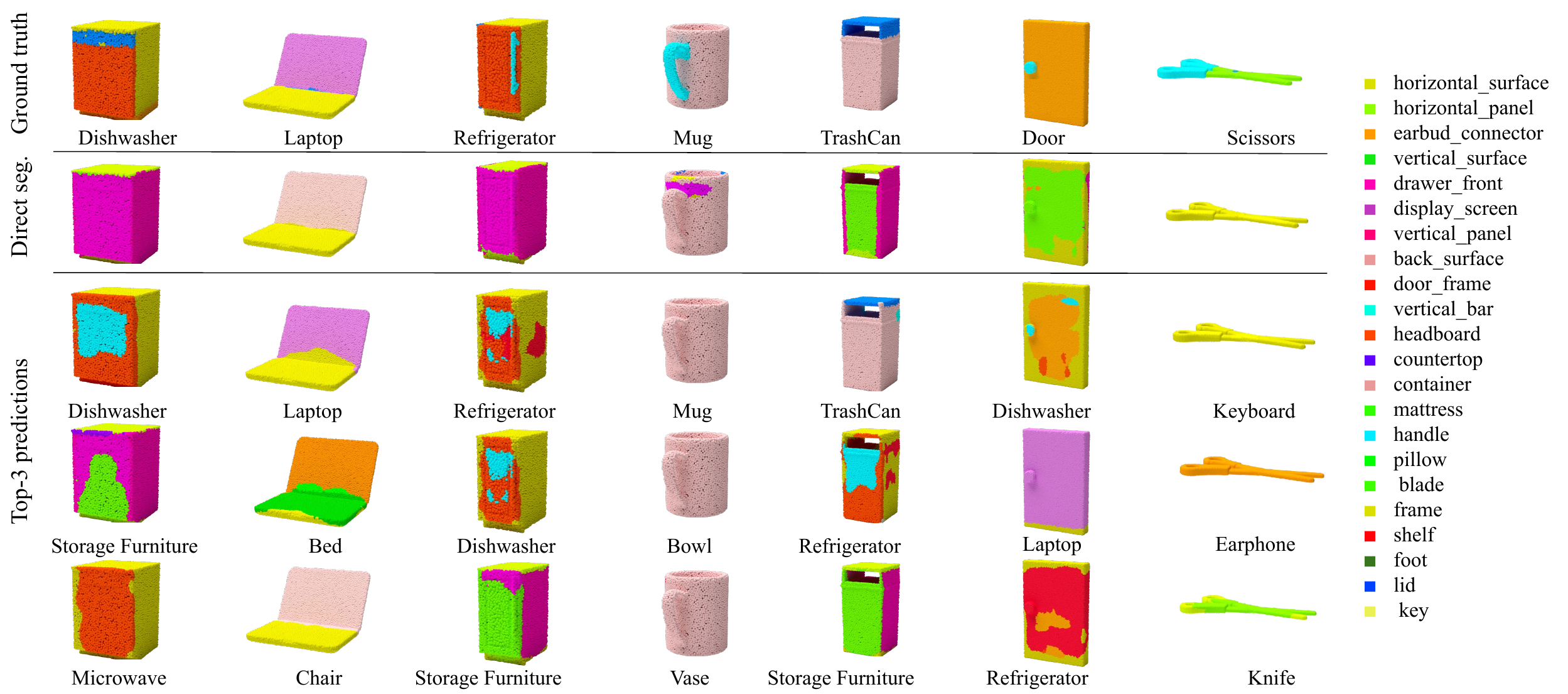}
   \caption{\textbf{Qualitative results.} Direct segmentation tends to segment an input point cloud to parts from seen objects with large geometric similarities. While this works for the objects from Container category, it fails in more complex objects that share similarity with Furniture while being composed of parts from other categories. 
   In contrast, DeCompositional Consensus builds an implicit understanding of what parts can occur together in different categories, and achieves meaningful segmentations for all object classes but Door and Scissors. 
   }
   \label{fig:qual}
\end{figure}

\myparagraph{Compositional zero-shot segmentation performance.}
Our method demonstrates remarkable performance gains on all unseen classes and achieves the best harmonic mean on compositional zero-shot segmentation in Table \ref{tab:basenumberseg}. We achieve a 50\% improvement over the direct segmentation demonstrating that the introduction of 
object class conditioned inference with DCC 
can improve compositional zero-shot segmentation in point cloud models. This improvement is observed most in unseen object classes that have large variations in parts from the seen object classes like Dishwasher (7.5$\times$), Laptop (2.5$\times$) and Refrigerator (28$\times$) as shown in Fig. \ref{fig:qual}. Unseen object classes that share large geometric and semantic similarities with respect to parts to seen object classes also have significant improvements. This includes improvements in Bowl (4\%), Mug (14\%), and TrashCan (1.4\%) that have very similar parts with seen Bottle and Vase. 

Comparing with zero-shot learning baselines, we observe that our method achieves the best performance in 6 out of 8 classes and establishes a state of the art in overall harmonic mean and unseen mIoU while achieving competitive seen IoU. 
PartPred \cite{attpred} learns to dynamically predict parts and generalizes to unseen objects that share part and geometric similarities with seen objects in the Container category but fails in other objects. As SPNet \cite{spnet} does not use any part information, it fails to generalize to unseen objects by projecting on word embeddings alone.
Compared to SPNet, CGE \cite{cge} performs much better as it uses the part prior and refines the word embeddings by using the dependency structure defined in the graph. Although being competitive on Bowl, Dishwasher, Mug and TrashCan, it performs much poorer on other unseen objects. 3D-PointCapsNet \cite{capsnet}, while conceptually engineered for part-whole relations, fails to generalize to unseen objects, likewise having very low performance on seen objects. We relate this performance to the capsules' inability to generalize without object prior as further shown in the supplementary material. 
Finally, PartPred DCC, achieves impressive performance on seen objects but fails on the unseen objects showing the importance of learning consensus over the features responsible for a hypothesis. We observe that while some methods have almost zero classification accuracy, they can still achieve some segmentation performance due to confusion with objects that share some parts with the ground truth object. All methods fail to generalize to challenging object classes Door and Scissors. We discuss that in qualitative analysis in Section \ref{sec:qual}.

\myparagraph{Zero-shot classification performance.} Our method also achieves significant gains on generalized zero-shot classification as seen in Table \ref{tab:basenumbercls}. DCC attains the best harmonic mean and unseen classification accuracy while maintaining a competitive seen performance. In fact, the best seen performance is achieved by PartPred DCC which extends our DCC score to a simple attribute (part) prediction model. This shows the power of enforcing consensus in different decisions of a model. Specifically, DCC is able to classify 6 out of the 8 unseen object classes with an outstanding accuracy.
SPNet is only able to classify Bowl with a low accuracy of 12\%. 
CGE is again a competitive baseline here. However, it is only able to receive reasonable classification scores on 4 of the 8 unseen object classes while maintaining a competitive seen class performance.

\subsection{Ablations}
We ablate our design choices and compare performance against different point cloud backbones.

\myparagraph{Optimization criteria.}
We ablate over the two optimization criteria for DCC in Table \ref{tab:dcablation}.
As seen from row a) that only training for $L_{DeComp}$ is unable to attain high performance as it can introduce bias against hard to predict parts to increase classification performance. Similarly, only training for $L_{Part}$ in row b) achieves low performance as the model is not optimized for the downstream classification task of predicting the consensus score. 
Row c) and d) combine both of these losses and see a big performance gain. In row c) we replace the predicted segmentation mask corresponding to the ground truth object class in HB with the ground truth segmentation mask.
Comparing row c) and d) in Table \ref{tab:dcablation}, we see that when we learn DCC score exclusively on the model's predicted segmentation instead of using ground truth segmentation mask, we see a large improvement in seen and unseen performance. We conjecture that the part scoring function $\mathcal{G}$ learns the segmentation network's limitations in this setting, \ie if a part is not predicted well by $\mathcal{F}$, $\mathcal{G}$ can look for cues from other parts.

{
\setlength{\tabcolsep}{2pt}
\renewcommand{\arraystretch}{1.3}
\begin{table}[t]
    \centering
    {
    \begin{tabular}{lccc|ccc|ccc}
    \multicolumn{4}{c}{\textbf{Hyperparameters}} & \multicolumn{3}{c}{\textbf{Classification}} & \multicolumn{3}{c}{\textbf{Segmentation}}\\
    & $L_{DeComp}$ & $L_{Part}$ & $Seg only$ &
    HM & S & U & HM & S & U \\
    \midrule
    
    a) & \checkmark& & \checkmark &  29.0 & 38.1 & 23.4 & 23.3 & 24.2 & 22.5 \\
    b) &  & \checkmark & &  14.9 & 34.8 & 9.4 & 24.8 & 22.1 & 28.3 \\
    c) & \checkmark & \checkmark &  & 52.6 & {54.4} & 50.9 & 39.1 & {35.9} & 42.9 \\
    d) & \checkmark & \checkmark & \checkmark & \textbf{72.8} & \textbf{76.6} & \textbf{69.3} & \textbf{45.1} & \textbf{40.9} & \textbf{50.2} \\
    \end{tabular}}
    
    \caption{\textbf{Ablating over $L_{DeComp}$ and $L_{Part}$}, we see that both the criterion complement each other to achieve the best performance.}
    \label{tab:dcablation}
\end{table}
}

    


\myparagraph{Comparing point cloud backbones.}
We compare point cloud backbones under Direct Segmentation and DCC in Table \ref{tab:bbzs}. 
We see that all models are unable to achieve competitive performance over unseen object classes with direct segmentation. In fact, ConvPoint \cite{convpoint} completely fails under this setting. The introduction of DCC to every backbone leads to a major increase in performance on the unseen object while being competitive over seen classes. This shows that our model can be readily extended to various families of point cloud backbones.
In Table \ref{tab:bboracle}, we compare the oracle segmentation performance over the ground truth object class when trained with and without our part prior optimization criterion ($L_{Seg}$ over $\mathcal{P}_o$ or over $\mathcal{P}$) . In absence of our criterion, we observe a large difference between the performance on seen and unseen object classes. We conjecture that the model overfits to seen object classes, limiting compositionality to unseen object classes. With our criterion, \eg PointNet segmentation network improves up to 19\% on unseen and 1\% on seen classes. 

{
\begin{table}[t]
    \begin{subtable}{0.49\linewidth}
    \resizebox{\linewidth}{!}
    {
    \begin{tabular}{l|ccc|ccc}
    \multirow{2}{*}{\bfseries Backbone}&
    \multicolumn{3}{c}{\textbf{Direct Seg}} &
    \multicolumn{3}{c}{\textbf{DCC}}\\
    & HM & S & U &
    HM & S & U \\
    \midrule
    PointNet \cite{qi2016pointnet} & 28.5 & 48.7 & 20.1 & 
    {35.2} & 38.0 & \textbf{32.7} \\
    DGCNN \cite{dgcnn} & \textbf{29.5} & \textbf{50.0} & \textbf{20.9} & 
    \textbf{36.2} & 45.1 & 30.2 \\
    ConvPoint \cite{convpoint} & 2.9 & 5.2 & 2.0 & 
    29.5 & 35.0 & 25.5 \\ 
    GDANet \cite{gdanet} & 28.7 & 47.7 & 20.5 & 
     33.5 & \textbf{46.2} & 26.4 \\ 
    \end{tabular}}
    \caption{\textbf{CZSL Segmentation}}
    \label{tab:bbzs}
    \end{subtable}
    \begin{subtable}{0.49\linewidth}
    \resizebox{\linewidth}{!}
    {
    \begin{tabular}{l|ccc|ccc }
    \multirow{2}{*}{\bfseries Backbone}& \multicolumn{3}{c}{\textbf{$L_{Seg}$ over $\mathcal{P}$}} & \multicolumn{3}{c}{\textbf{$L_{Seg}$ over $\mathcal{P}_o$}} \\
    & HM & S & U & HM & S & U \\
    \midrule
    PointNet \cite{qi2016pointnet} & 43.2 & 51.7 & 37.1 & {47.9} & 52.8 & {43.8} \\
    DGCNN \cite{dgcnn} & \textbf{44.6} & 52.4 & \textbf{38.8} & \textbf{50.0} & \textbf{55.0} & \textbf{46.3} \\
    ConvPoint \cite{convpoint} & 29.1 & 28.7 & 29.5 & 43.5 & 42.4 & 43.0\\
    GDANet \cite{gdanet} &  43.4 & \textbf{53.1} & 36.8 & 48.0 & 53.7 & 43.4 \\ 
    \end{tabular}}
    \caption{\textbf{Oracle Performance}}
    \label{tab:bboracle}
    \end{subtable}
    
    \caption{\textbf{Backbone ablation}. (a) We see DCC results in a large improvement compared to direct segmentation across all ablated point cloud models (b) We further see that our Part Prior optimization criterion greatly benefits all backbones under oracle evaluation especially on unseen objects.}\label{tab:backbone}

\end{table}
}

\subsection{Qualitative and Model Limitation}
\label{sec:qual}
In Figure \ref{fig:qual}, we show some qualitative results for direct segmentation versus top-3 results of our model across unseen objects. We further validate our results from Table \ref{tab:basenumberseg}, and see that for the easy object Mug, the direct segmentation can give a meaningful result. However, it fails for other relatively harder objects, which can be attributed to the lack of affordance, \ie how an agent interacts with an object.
Our method, despite not having access to affordances, builds an implicit understanding of what parts occur in each object category and is thus able to learn a reasonable consensus score. This brings about an increased generalization and meaningful results for all objects. 
We see a correct prediction for even Dishwasher and Refrigerator, which are closer geometrically to Furniture than Microwave, their closest functional seen object. However, although less, DCC also suffers from lack of affordances. For example, among the top-3 result for a Laptop in Figure \ref{fig:qual} is a Chair which shares geometrical similarity to an open Laptop. This indicates an upper bound to the performance that can be achieved from visual data alone \cite{aff1,aff2}. An affordance prior can help address this limitation for part-object relations.

\begin{figure*}[t]
\centering
\includegraphics[width=0.8\linewidth]{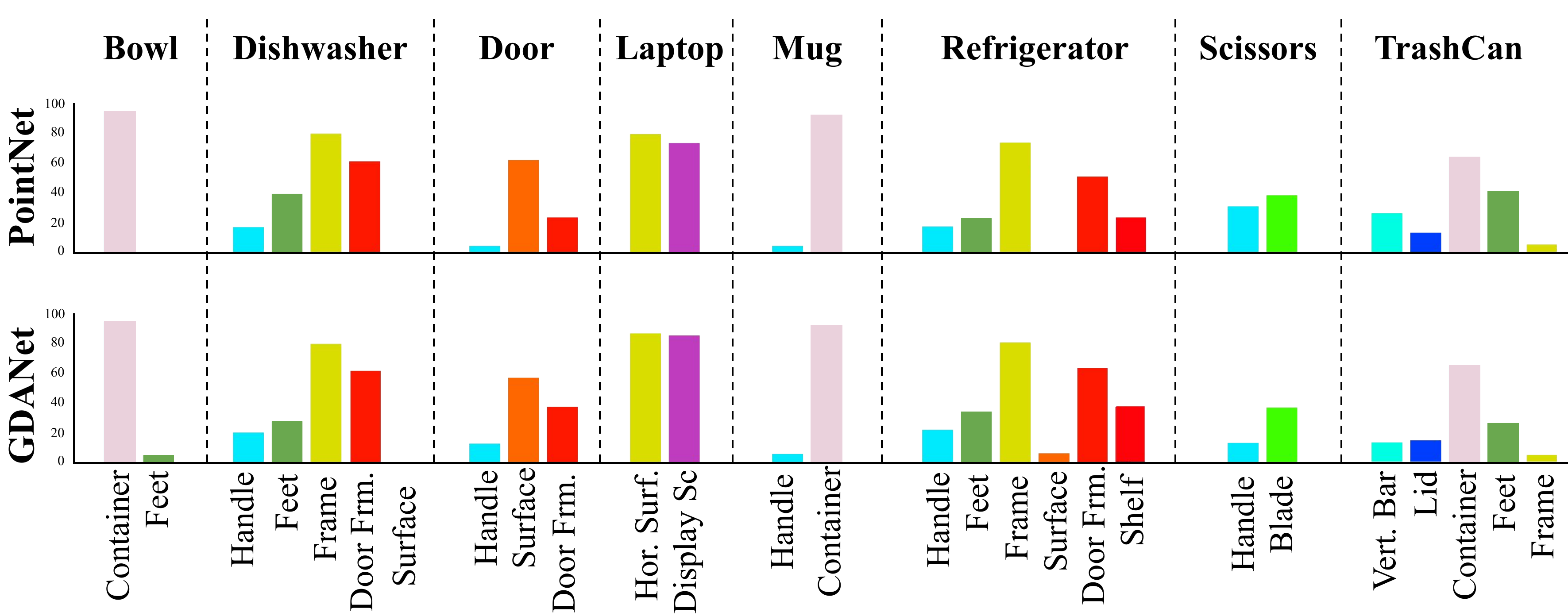}
\caption{\textbf{Error plots.} We find that PointNet\cite{qi2016pointnet} is comparable in mIoU to a much newer model, GDANet\cite{gdanet}, across unseen objects.}
\label{fig:barplot}
\end{figure*}

Another aspect that limits our model performance is the generalization limitations of point cloud backbones. 
In Figure \ref{fig:barplot}, we compare the object prior per part performance on the unseen objects between PointNet \cite{qi2016pointnet} and GDANet \cite{gdanet}, which were released five years apart. 
A surprising insight we observe is that years of progress in point cloud processing, while making a significant advance on seen object classes, does not translate to improvement on unseen object classes. 
We see that there is no clear consensus on which model is better for unseen object class generalization. 
Even using the right part prior, some parts are unlikely to be segmented in unseen classes. An example of this is handle, which is unable to be reasonably segmented for Mug,  Dishwasher, and Refrigerator. A more extreme case of this is observed in Door and Scissors, where the segmentation fails completely 
as shown in last two columns of Figure \ref{fig:qual}. 
These objects have a large variation with respect to parts from the seen objects in scale, the number of instances (two blades in Scissors vs one in Knife), and orientation.

\section{Conclusion and Future work}
We introduce 3D-CZSL as a joint compositional zero-shot segmentation and generalized zero-shot classification task. We provide a structured study into zero-shot compositionality through a novel benchmark on the proposed C-PartNet dataset and show that previous models do not generalize beyond the training object classes. Towards this, our novel approach, DeCompositional Consensus, maximizes the agreement between a segmentation hypothesis and its parts when taken apart, and sets a new SOTA. We also show that while there has been a lot of progress in part segmentation in a supervised setting, simple models like PointNet are still competitive in unseen object classes, arguably because the current research has not focused on this task.
There are several future directions that can stem from this work, including introducing affordance priors and extension of the capsules paradigm for part reasoning on unseen object classes. We also hope to inspire future research into compositional point cloud models.

\clearpage
%
%
\bibliographystyle{splncs04}
\bibliography{egbib}
\end{document}